# Generating Individual Travel Diaries Using Large Language Models Informed by Census and Land-Use Data

**Sepehr Golrokh Amin[a], Devin Rhoads[a], Fatemeh Fakhrmoosavi, Ph.D.*[a], Nicholas E. Lownes, Ph.D.[a], John N. Ivan, Ph.D.[a]**

[a] School of Civil and Environmental Engineering, University of Connecticut, United States

* Corresponding author
E-mail: moosavi@uconn.edu

## Abstract

This study introduces a Large Language Model (LLM) scheme for generating key attributes of travel diaries in agent-based transportation models, including purpose, mode and distance, to assess the underlying viability of LLMs for activity generation tasks. While traditional approaches rely on large quantities of proprietary household travel surveys, our method generates personas stochastically from open-source American Community Survey (ACS) and Smart Location Database (SLD) data, then synthesizes diaries through direct prompting. Our study features a novel one-to-cohort realism score: a composite of four metrics (Trip Count Score, Interval Score, Purpose Score, and Mode Score) validated against the Connecticut Statewide Transportation Study (CSTS) diaries, matched across demographic variables. Our validation utilizes Jensen-Shannon Divergence to measure distributional similarities between generated and real diaries. When compared to diaries generated with classical methods (Negative Binomial for trip generation; Multinomial Logit for mode/purpose) calibrated on the validation set, LLM-generated diaries achieve comparable overall realism (LLM mean: 0.692 vs. 0.628). The LLM excels in determining trip purpose, and its trip mode predictions demonstrate greater consistency (a narrower Realism Score distribution). Meanwhile, classical models lead to better numerical estimates of trip count and activity duration. Aggregate validation confirms the LLM's statistical representativeness (LLM mean: 0.779 vs. 0.706), demonstrating LLM's zero-shot viability and establishing a quantifiable metric of diary realism for future synthetic diary evaluation systems.

## 1. Introduction

Transportation planning directly affects daily life, shaping everything from traffic congestion to environmental health. At its core, this field relies on understanding and predicting individual travel behaviors, which are notoriously complex and heterogeneous. Income, household size, mobility options such as ride-hailing services, transit availability, and car ownership all directly influence travel patterns within communities (Fakhrmoosavi et al., 2024; Paithankar et al., 2024; Püschel et al., 2024). Agent-Based Models (ABMs) are widely used to simulate individual behavior (Fakhrmoosavi et al., 2025), but they come with two major challenges. First, they often need detailed survey data for calibration, which can be difficult or expensive to obtain (Mori et al., 2026). Second, their mathematical structures tend to be rigid, making it hard to capture the subtle, context-dependent ways that people actually make decisions (Liu et al., 2025; Wang et al., 2024). To address these issues, we introduce a novel approach that uses the reasoning and language capabilities of Large Language Models (LLMs). Our main question is whether LLMs, operating in a zero-shot setting, without any prior training on travel survey data and guided only by publicly available census and land-use data, can produce daily travel diaries that are both realistic and interpretable. The approach presented in this study can reduce the heavy data requirements of traditional ABMs while improving how well they reflect real human behavior.

While traditional four-step models (trip generation, distribution, mode choice, and assignment) have long served as the backbone of transportation planning, they often rely on aggregated data and treat each modeling step in isolation. As a result, these models struggle to accurately represent the nuanced, individual-level decision-making influenced by factors such as income, disability status, or household dynamics. Ignoring these complexities may compromise model accuracy and limit its practical value for infrastructure investments and its ability to positively influence daily quality of life (Mladenovic and Trifunovic, 2014; Hörl and Balać, 2020). ABMs addressed some of these limitations by modeling travel as sequences of linked activities, but they require extensive data input and are computationally intensive, making them difficult to implement or maintain on a regular basis (Rezvany et al., 2024).

Machine learning (ML) has emerged as a promising alternative, using algorithms like Random Forests and Neural Networks to capture the complex, nonlinear relationships that traditional approaches often lack (Rostami et al., 2024). For instance, ML has outperformed gravity models in trip distribution when paired with New York City (NYC) American Community Survey (ACS) and social media data (Pourebrahim et al., 2019) and has shown a robust ability to improve trip generation accuracy. Notably, under simulated conditions with high data obfuscation (over 90% of trips missing), a well-specified model could recover between 81% and 95.4% of the ground-truth trips, indicating high accuracy with the potential for slight over- or under-estimation (Guan et al., 2025). In freight forecasting, Support Vector and Gaussian Process Regression have reduced error by up to 30% compared to standard regression (Lim et al., 2022).

Despite these gains, most research in this area remains narrowly focused. Some studies used Recurrent Neural Networks (RNNs), particularly Long Short-Term Memory models (LSTMs), to forecast short-term travel demand, such as taxi ride volumes in NYC, based on time-sensitive factors like weather and day of the week (Kim et al., 2020). Graph Neural Networks (GNNs) have also been employed to enhance trip generation and station placement decisions in bike-sharing systems (Liang et al., 2023). These models improve accuracy and interpretability but are rarely integrated across multiple stages of the travel demand modeling process.

Recently, researchers have begun experimenting with quantum-inspired algorithms for optimization in transportation. While initial work largely focused on theoretical reviews of Quantum Neural Networks (QNNs) and Quantum Convolutional Neural Networks (QCNNs) (Zhuang et al., 2024), practical applications are beginning to emerge. For example, studies have used hybrid quantum-classical models for traffic prediction (Zhang et al., 2020) and Q-learning reinforcement learning with Gaussian and Poisson reward models to optimize mode choices under variable traffic densities (Taş et al., 2025). While these developments show promise in improving optimization performance, their scalability and readiness for integration into comprehensive transportation modeling frameworks remain limited.

Several other advanced modeling methods have enabled researchers to better predict destination and mode choices. For example, the Genetic Algorithm was used to simultaneously estimate parameters for different travel decisions, like destination and mode choice, improving accuracy despite their computational complexity (Bernardin et al., 2015). Building on the developments in advanced modeling techniques, one study compared seven ML methods using repeated k-fold cross-validation and found Random Forest to be most accurate for predicting travel modes, although the influence of specific factors varied across methods (Hagenauer and Helbich, 2017). Together, these studies highlight a growing adoption of advanced computational techniques in transportation modeling.

Advanced statistical and machine-learning methods can provide strong predictive performance for individual travel-demand components, including trip generation and mode choice (Hagenauer and Helbich, 2017; Püschel et al., 2024). However, these approaches generally require labeled travel observations for model estimation, calibration, tuning, and local validation. The present study examines a different use case: whether a general-purpose LLM can generate plausible daily travel diaries without study-specific training or calibration using household travel-survey records. Instead, the LLM receives synthetic personas and land-use contexts constructed from publicly available ACS and SLD data. Llama 3 was not fine-tuned or calibrated using the Connecticut HTS records employed for validation. The objective is not to replace calibrated statistical or machine-learning models where sufficiently detailed local data are available, but to evaluate a complementary diary-generation approach that may be useful in settings where such data are limited or unavailable.

To evaluate the realism of LLM-generated diaries, we utilize a validation strategy that compares them against two sources. The first source is the ground truth dataset: the 2016-2017 Connecticut Statewide Transportation Study (CSTS). The second is a set of diaries constructed using traditional statistical models calibrated on the CSTS itself. The latter provides a direct assessment of LLM performance against a benchmark constructed by numerical models explicitly trained on the CSTS dataset itself. The primary validation metric is a one-to-cohort analysis, where each synthetic diary is compared against a cohort of its direct real-world peers, matched on six key demographic variables. This is supplemented by an aggregate-level analysis using Jensen-Shannon Divergence (JSD) to assess how closely the pooled distributions of generated travel attributes match the corresponding HTS population distributions (Wang et al., 2024).

This study offers two distinct contributions that directly address these challenges. First, we introduce a locally implemented diary-generation method that does not require household travel-survey data to train or fine-tune the LLM. Persona synthesis uses publicly available ACS and SLD data in a zero-shot way, while the Connecticut household travel survey is used only for validation and calibration of the comparison benchmark. The framework is executed locally without external model APIs. Second, the established one-to-cohort analysis assigns a holistic score to the realism of zero-shot generations beyond traditional aggregate comparisons. This provides a nuanced, person-level assessment of behavioral realism by comparing synthetic agents against direct demographic peers, setting a stricter standard for evaluating generative models. By presenting a fully implemented and validated system, our work provides a practical application that moves beyond previous conceptual frameworks.

The paper proceeds as follows: we first describe the datasets for persona generation and validation; next, we present our two-stage methodology for generating synthetic travel diaries and the one-to-cohort validation strategy; we then report and discuss experimental results, comparing the LLM to a classical modeling benchmark. Finally, we summarize key findings, discuss limitations, and outline future research directions.

## 2. Data description

This study utilizes three primary sources of data. The 2015 ACS and associated land-use data serve as the inputs for generating our representative personas. Specifically, this study uses the 5-year summary file data that provides statistical distributions of variables at the census block group level, allowing for persona synthesis to be based in local context. We chose the 2015 ACS as the closest timeframe to our 2016–2017 ground truth survey. This was necessary as large-scale travel surveys are costly and infrequent. Our methodology itself is flexible and applicable to newer data. Also supplementary to the LLM prompt generation is the Environmental Protection Agency's (EPA) SLD, which includes up to 90 (dependent on location) attributes for measuring land use and physical locational efficiency, including metrics such as housing and employment density and distance to nearest transit. Both the ACS and SLD provide context within prompt generation that is expected to change travel behavior; for example, a person living near transit is more likely to choose transit as their mode when compared to a person living in a low-density block group with poor transit access.

Comparatively, the CSTS dataset serves as the ground truth for validating the realism of the generated travel diaries in this study. The CSTS dataset includes rich, multi-level data at the household, person, and trip levels. These data, which is described in detail below, helps explain differences in travel behavior. Variables like home ownership and housing type are often linked to factors such as vehicle access, travel frequency, and preferred modes. For example, people living in single-family homes may be more likely to own multiple vehicles, while those in multi-unit buildings may rely more on transit or walking. These characteristics, as shown in Table 1 and Figure 1, are included in the modeling framework to capture variation in travel patterns across different types of households.

**Table 1**
Descriptive statistics of household- and trip-level characteristics.

| Variable | Unit | Min | Median | Mean | Max | SD |
|---|---|---|---|---|---|---|
| *Household Characteristics (sample=8,403*)* | | | | | | |
| Household size | Persons | 1 | 2 | 2.47 | 9 | 1.36 |
| Number of adults | Persons | 1 | 2 | 1.93 | 6 | 0.86 |
| Number of children | Persons | 0 | 0 | 0.54 | 6 | 0.94 |
| Number of workers | Persons | 0 | 1 | 1.21 | 6 | 0.95 |
| Number of drivers | Persons | 0 | 2 | 1.74 | 6 | 0.87 |
| Number of vehicles | Vehicles | 0 | 2 | 1.78 | 9 | 1.08 |
| Number of bicycles | Bicycles | 0 | 1 | 1.15 | 5 | 1.46 |
| *Trip Characteristics (sample=58,534)* * | | | | | | |
| Trip Duration | Minutes | 5.00 | 15.00 | 16.47 | 70.00 | 11.61 |
| Trip Distance | Miles | 0.00 | 3.00 | 5.11 | 26.75 | 5.60 |
| Number of Travelers | Persons | 1.00 | 1.00 | 1.68 | 5.00 | 0.99 |
| *Household Income Category (sample=8,403*)* | | Percentage of Households | | | | |
| Less than $25k | Indicator | 15.4% | | | | |
| $25k–$49,999 | Indicator | 16.3% | | | | |
| $50k–$74,999 | Indicator | 13.6% | | | | |
| $75k–$99,999 | Indicator | 11.4% | | | | |
| $100k or more | Indicator | 27.2% | | | | |
| Prefer not to answer | Indicator | 16.1% | | | | |

Min = minimum, Max = maximum, SD = standard deviation.
*: All statistics are calculated using survey weights to reflect the population.

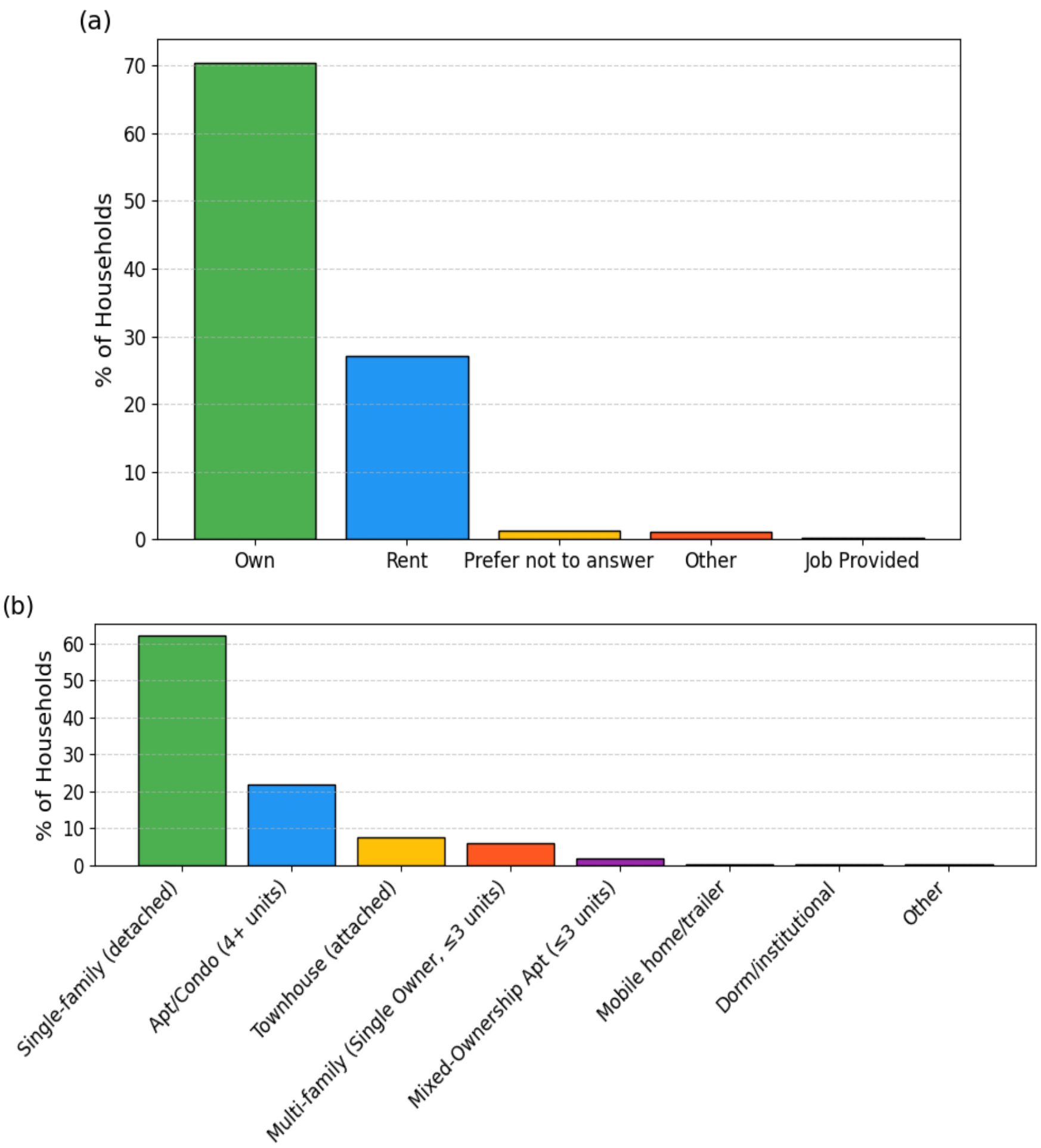

**Fig. 1.** Household housing characteristics; (a) occupancy status (b) residence type.

A manual inspection of the trip data revealed records that contained extreme, infeasible values that would disproportionately skew analyses. For example, the raw dataset contained single trips with reports exceeding 18 hours and over 3,400 miles. Given that such records are likely data entry errors, trips were removed from the dataset if their distance, duration, or number of travelers fell more than 3 times the Interquartile Range (IQR) above the third quartile or below the first quartile. This filtering process reduced the sample from 65,103 to 58,534 trips, removing approximately 10.1% of the sample.

Note that trip distances of 0.00 miles (Table 1) persist, as they were not statistical outliers. Such values reflect valid, very short trips (like walking to a neighboring building) that were not removed by the filter. In contrast, the minimum observed trip duration is 5 minutes. While <5-minute trips are plausible, they are absent in the final dataset, likely as an artifact of survey precision (e.g., rounding durations to 5-minute intervals) combined with the outlier removal process. After filtering for extreme outliers, trip-level characteristics indicate that on average, trips last about 17 minutes and cover 5.11 miles, though the high standard deviations reflect a wide range of travel behavior. Most trips involve one person, but some include up to 5 travelers, capturing both everyday and group travel.

Figure 2. Distribution of trip departures across 24 contiguous hourly intervals from 3:00 AM to 3:00 AM the following day. Each bar represents a left-closed, right-open interval, [t, t + 1

hour), between adjacent horizontal-axis labels. Travel peaks appear in the morning from 7:00 AM-9:00 AM and in the afternoon from 3:00 PM-6:00 PM, matching typical commute times for work and school schedules. These patterns help explain daily travel routines and support models capable of accounting for when and why people choose to travel. Figure 3 shows that most trips use household vehicles, with fewer involving walking, public transit, or other modes, which is an important input for travel behavior modeling.

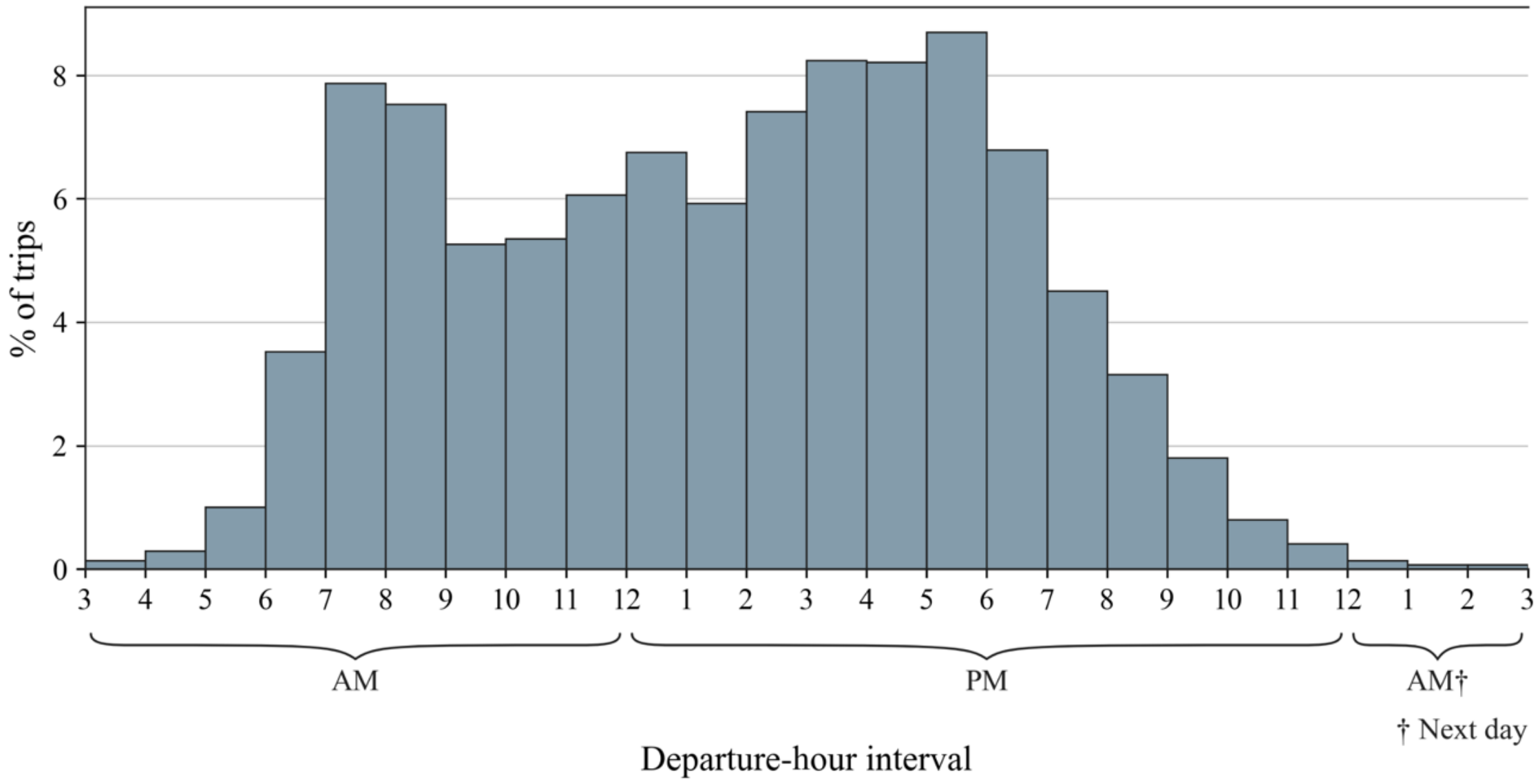

**Fig. 2.** The distribution of trip departure times over a 24-hour period.

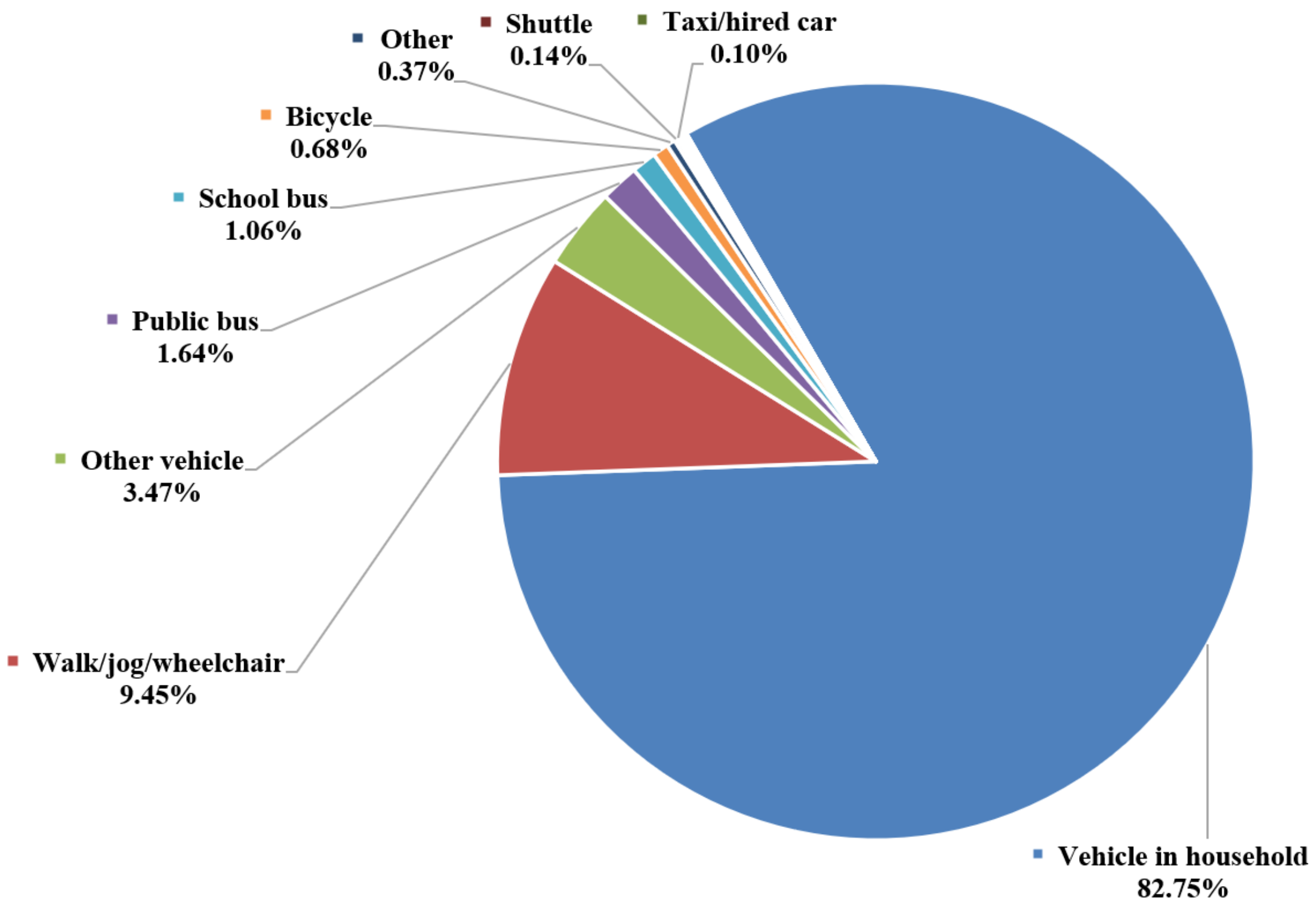

**Fig. 3.** The transportation mode distribution (modes <0.10% not labelled).

## 3. Methodology

To address the data dependency and behavioral limitations of traditional ABMs, a novel framework for generating realistic and interpretable daily travel diaries was developed. This approach leverages the reasoning capabilities of a locally run LLM to create a synthetic, census-based persona for a given geographic area and then directly generate its travel diary. This method is grounded in publicly available census and land-use data and does not require proprietary HTS data for model training or fine-tuning. In other words, the model's geographic context is not derived from the location's name, but from the prompt itself, which details the specific demographic and land-use characteristics of a given local block group. This section details the framework's architecture, the persona synthesis and generation process, and the rigorous validation strategy used to assess the realism of its outputs.

### *3.1. Framework overview*

The core of our methodology is a two-stage process to generate a travel diary for a single, synthetic individual. While this process shares similarities with population synthesis, our approach differs in both goal and scope. Traditional population synthesis aims to create a complete and internally consistent synthetic population for a region (often using techniques like iterative proportional fitting), whereas our method focuses on generating a single, plausible but independent persona for each block group.

In the first stage, the framework generates a realistic virtual persona by randomly sampling demographic and economic data from a specific neighborhood (census block group).This ensures the persona is typical for that area but is not just a generic "average" person. Second, the framework feeds the LLM details about that persona's neighborhood, like land use and population characteristics to guide it in generating a daily travel diary. A critical part of the framework design is that the LLM is never given the actual name of the town or state. It has to infer the "sense of place" purely from the statistical clues in the prompt.

Note that the objective of this study is not to generate a complete simulated population, but rather to evaluate the LLM's ability to perform across a wide range of local contexts without prior fine-tuning. This enables a rigorous, real-world assessment of its zero-shot reasoning capabilities. Accordingly, the framework is not inherently tied to Connecticut and can be applied to other regions using the same general data-processing and prompting workflow. However, applying the framework to a new region requires more than replacing the study-area ACS and SLD inputs with local data. Classification thresholds, persona-synthesis parameters, land-use descriptions, and travel-mode and trip-purpose categories may also need to be adjusted to reflect local conditions. The Connecticut case study demonstrates the framework's feasibility in one geographic setting, while its performance in other regions remains to be evaluated. The primary contribution of this work lies not in developing a new population synthesis technique, but in demonstrating the use of persona synthesis as a sampling strategy for generating diverse and contextually grounded prompts for LLMs. Figure 4 shows the LLM diary generation workflow.

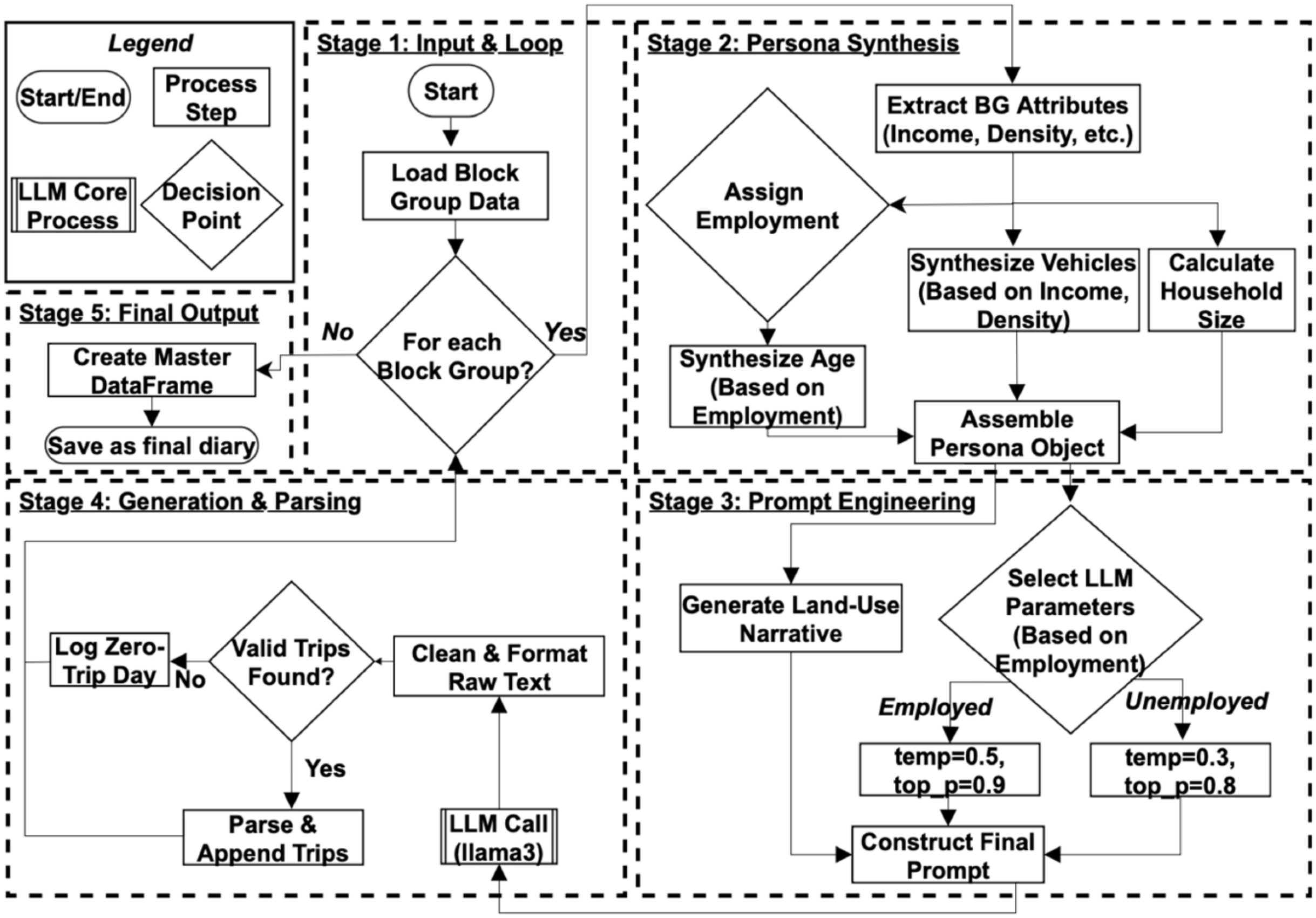

**Fig. 4.** The LLM diary generation workflow.

As illustrated in Figure 4, the framework employs a sequential workflow comprising input preparation, stochastic persona synthesis, prompt construction, and direct LLM diary generation to translate aggregate census and land-use data into individualized travel behaviors. This decoupled design ensures that every synthetic trajectory is grounded in a demographically consistent agent prior to behavioral prompting. Executed across all Census Block Groups, the pipeline utilizes the 2016–2017 CSTS dataset exclusively for downstream validation, preventing data leakage. Detailed implementations of these stages follow in the subsequent sections.

### *3.2. Technical implementation*

The core of the system is Meta's Llama 3, which was accessed through Ollama and configured to run locally without relying on third-party APIs (Llama 3 model card - Github, 2024; Meta, 2025). Llama 3 is a general, pre-trained foundation model trained on publicly available text and code, not on any specialized travel survey data from Connecticut or elsewhere. This property is fundamental to our zero-shot approach, as the model's ability to generate plausible travel diaries stems from its emergent reasoning capabilities rather than prior exposure to specific behavioral datasets.

Llama 3 was selected for its strong performance in zero-shot reasoning tasks, where it has shown superior results in human-mobility prediction compared to models such as Mistral, Llama 2, GPT-3.5, and GPT-4o. This setup not only protects sensitive data by keeping all processing local but also provides a compelling test of the model's reasoning abilities under zero-shot conditions. All data manipulation is handled in Python while Ollama manages local model interaction (Beneduce et al., 2025).

*3.3. Persona and context generation*

For each block group, two primary inputs are generated based on SLD and ACS data, linked by the census block group "GEOID". The input data for this process includes block-group-level statistics such as employment rate, median household income, intersection density, total population, and total household count (from the SLD).

1- The Synthetic Persona: Instead of using simple aggregate measures, we create the persona through a multi-step stochastic process designed to generate a plausible individual representative of the block group's statistical profile. This is achieved by probabilistically assigning key demographic attributes in a specific sequence:

- Employment Status: Employment status is modeled as a Bernoulli random variable with probability (p) equal to the block-group employment rate. A uniform random draw, $u$, from [0,1) assigns the persona to the employed category when $u < p$ and to the alternative category otherwise.
  Age Bracket: Age bracket is represented as an ordered categorical attribute and sampled from a predefined, nonuniform probability profile conditional on the assigned employment status. The profile assigns relatively greater probability to prime working-age categories for employed personas and relatively greater probability to younger and older categories for the alternative group. This directional structure is consistent with age-specific employment patterns reported by the U.S. Bureau of Labor Statistics (2015).
- Household Vehicle Count: Household vehicle count is also represented as an ordered categorical attribute and sampled from a predefined, nonuniform probability profile. The profile shifts toward higher vehicle-count categories as block-group income increases and toward lower vehicle-count categories as intersection density increases. These directional relationships are consistent with empirical evidence of positive associations between income and vehicle ownership and lower vehicle ownership in denser urban environments (Dargay and Gately, 1999; Giuliano and Dargay, 2006; Holian, 2020). The adjusted probability profile is normalized before sampling.
- Household Size: This attribute is calculated deterministically, not stochastically. It is derived from the block group's average by dividing the total population by the number of households (a variable sourced from the SLD) and rounding the result to the nearest integer to ensure realism.

Weighted random sampling was used because age bracket and household vehicle count are discrete attributes with nonuniform category probabilities. Sampling from a specified probability mass function is a standard method for generating discrete random variables (Devroye, 2006). The predefined probability profiles and adjustment rules are applied uniformly across geographic units, while the resulting probabilities vary according to the block-group characteristics used as inputs. These procedures are intended to generate heterogeneous personas for evaluating zero-shot travel diary generation and do not constitute a calibrated population-synthesis model. This entire synthesis process is implemented in Python, utilizing functions from the random library, such as "random" and "choices", to perform the probabilistic assignments. By explicitly defining the conditional probabilities and adjustments, the synthesis process, while stochastic, is formally modeled and fully reproducible, directly addressing the need for a more rigorous and less narrative-driven methodology.

2- The Land-Use Context: A string describing the physical environment, based on SLD variables such as population density, employment mix, and transit accessibility.

*3.4. Direct diary generation*

The diary-generation pipeline includes procedures for handling model-call and output-format failures. If the local model call raises an exception, the routine makes up to three total attempts, with a one-second pause between unsuccessful attempts. Once a response is received, the returned text is screened line by line and parsed using a standard CSV reader. A record is retained only if it contains exactly the five expected fields: start time, end time, trip purpose, travel mode, and travel distance. Records with fewer or more than five fields are discarded. This procedure verifies the structural format of each record but does not independently validate the semantic content of each individual field.

If all model-call attempts fail, the response is empty, or none of the returned records passes the structural check, the pipeline retains the persona using a placeholder record with “N/A” for start time, end time, and travel mode, “Generation failed” for trip purpose, and zero travel distance. This placeholder indicates that no structurally valid trip record was available and prevents the persona from being omitted from output. A valid non-travel day is represented by an explicit zero-trip record and labeled “valid zero-trip,” whereas the fallback record is labeled “model failure” in a separate status field. The overall diary-generation workflow is presented in Figure 4.

With a fully defined synthetic persona, the final diary is generated via a single, comprehensive prompt passed to the Llama 3 model. This prompt instructs the LLM to "act as" the assigned persona and generate a plausible weekday travel diary. The prompt includes several critical instructions to enhance realism and ensure analyzable outputs:

- It provides the LLM with complete context, including the persona's full demographic profile and the land-use characteristics of their home environment.
- It constrains the LLM's output for trip purposes and travel mode to a predefined list of standardized categories to ensure compatibility with the HTS data dictionary.
- The framework dynamically adjusts the LLM’s decoding parameters based on the persona’s employment status to better reflect behavioral differences. Specifically, two techniques temperature and top_p are used to control the generative process. These methods have been shown to effectively generate outputs whose statistical properties closely match those of human activity patterns, particularly with top_p values around 0.9. Temperature serves as a complementary parameter to fine-tune output randomness, balancing quality and diversity; lower values improve coherence but may reduce variation if set too low (Ackley et al., 1985; Holtzman et al., 2020).

  Based on the hypothesis that employed individuals exhibit more varied travel behavior, parameters were heuristically selected to balance plausibility and diversity. These initial parameters were selected based on established practices in the literature and confirmed through a qualitative review of initial outputs for plausibility. Employed personas were assigned a top_p of 0.9 and a temperature of 0.5 to encourage greater diary diversity. Conversely, unemployed personas received a lower top_p of 0.8 and a temperature of 0.3 to produce more consistent, routine-based diaries. This approach enables the framework to generate diaries that embody distinct, justifiable assumptions about behavioral randomness.

Because LLM decoding uses probabilistic sampling, repeated model calls can produce different individual travel diaries. Temperature and top_p control the degree of variation in token sampling, while the random seed affects the sampling sequence. A fixed seed can improve repeatability but does not necessarily guarantee identical outputs across different model versions or software environments. Such variation at the individual-diary level is expected because daily travel behavior itself is stochastic; the more relevant consideration for synthetic travel diary generation is whether the resulting aggregate travel patterns remain stable across repeated executions.

To assess run-to-run stability, the complete pipeline was executed seven times using the same block-group input file, Llama 3 model, prompt construction procedure, and decoding settings. Because both persona synthesis and model decoding remained stochastic across executions, this experiment evaluates the stability of the complete diary-generation pipeline rather than exact replication of individual diaries. Aggregate stability was assessed across all $\binom{7}{2} = 21$ unique pairs of runs using one minus the JSD for the trip-count, trip-purpose, travel-mode, and activity-interval distributions. The resulting stability measures are reported in Section 5.4, and the JSD calculation is defined in Section 4.1.

## 4. Validation strategy

To assess the realism of LLM-generated diaries, we developed a validation framework comparing model outputs to both real-world survey data and a traditional travel demand benchmark. The validation set, constructed from CSTS data, comprises chronological, event-based survey diaries with person-records containing mapped demographics (e.g., age, income) and sequential “Activity” and “Trip” events.

### *4.1. One-to-Cohort validation*

The primary validation method in this study is a highly granular "one-to-cohort" analysis. For each individual synthetic diary (from both the LLM and the classical benchmark), a corresponding cohort of direct peers within the real HTS dataset is identified. This matching is performed using six demographic variables: age bracket, employment status, household vehicle count, income level, GEOID, and household size. A "Realism Score" is then calculated for each synthetic diary by comparing its travel patterns (e.g., number of trips, mode distribution, purpose distribution) against the average behavior of its real-world cohort. This score is a composite metric on a scale from 0 to 1, where 0 indicates a complete mismatch and 1 a perfect match; conceptually, it quantifies the extent to which a synthetic diary successfully captures the dominant travel patterns of its real-world peers. To avoid calculating person-level realism scores using very small comparison cohorts, a prespecified hierarchical fallback procedure is applied when fewer than 10 unique HTS individuals satisfy the initial matching criteria. As shown in Figure 5, the matching process begins with the full set of six variables. If fewer than 10 unique individuals are identified, GEOID is removed to form a five-variable match, followed by household size to form a four-variable match. If the cohort remains below the minimum threshold, matching is broadened to age bracket and employment status only. Finally, if fewer than 10 individuals are identified using these criteria, the complete HTS sample is used as the reference population. This procedure retains the most detailed feasible demographic matching while avoiding comparisons based on very small reference cohorts. These cohort-matching levels are referred to as "Hyper Strict," "Ultra-Strict," "Strict," and "Broad" in the One-to-Cohort Validation Performance Section. The Overall Realism Score is a composite of four components.

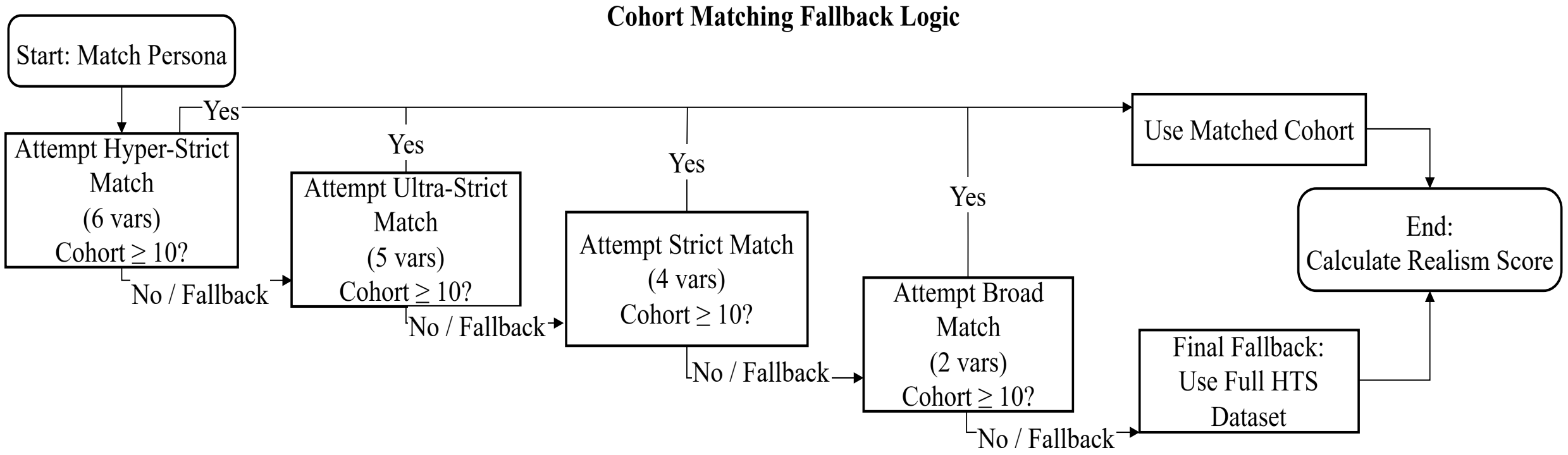

**Fig. 5.** Cohort matching fallback logic workflow.

To measure the similarity for trip purpose, activity interval, and travel mode, this study uses the JSD. JSD is a method used to quantify the similarity between two discrete probability distributions and was selected due to its suitability for comparing categorical data and ability to handle zero-value categories (Sason, 2022); if a generated diary includes a trip purpose not found in its respective cohort, JSD will still produce a valid score. While uncommon in transportation literature, JSD was used before (Wang et al., 2024) to similarly compare distributions of generated and real-world trajectories. However, we acknowledge its limitations; JSD can be sensitive to the number of bins chosen for continuous data (such as activity intervals) and may be less stable with sparse distributions generated from a single diary. Future work could explore bootstrapped confidence intervals to strengthen the measure of distributional similarity (Efron and Narasimhan, 2020).

For these comparisons, the distribution of the HTS cohort is denoted as P, and the distribution from the single generated diary is denoted as Q. Both distributions consist of discrete outcomes represented by the variable, $x$. For example, when calculating Mode Score (Equation 5), $x$ would represent a specific travel mode (i.e., "Bicycle"). The JSD between the generated distribution, $Q$, and cohort distribution, $P$, is computed using the Jensen Shannon function of the SciPy module (Virtanen et al., 2020). Because this function returns the square root of the Jensen–Shannon divergence, the returned value was squared to obtain the base-2 divergence. Each distributional similarity score was then calculated as one minus this divergence. The JSD is defined as:

$$JSD(P||Q) = \frac{1}{2}D_{KL}(P||M) + \frac{1}{2}D_{KL}(Q||M) \quad (1)$$

In Equation (1), $D_{KL}$ is the Kullback-Leibler divergence, calculated as $D_{KL}(A||B) = \sum_{x \in X} A(x) \log_2(\frac{A(x)}{B(x)})$, and $M = \frac{1}{2}(P + Q)$ represents the midpoint distribution between ground-truth and generated sets, $P$ and $Q$. The four component scores that comprise the Overall Realism Score are calculated as follows:

1. Trip Count Score: Measures how close the number of trips in the synthetic diary, $n_{generated}$, is to the average number of trips in the cohort, $\mu_{cohort}$.

$$\text{Trip Count Score} = 1 - \min\left(1, \frac{|n_{generated} - \mu_{cohort}|}{\mu_{cohort}}\right) \quad (2)$$

2. Purpose Distribution Score: Measures the similarity between the generated diary's purpose distribution and the cohort's purpose distribution using JSD.

$$\text{Purpose Distribution Score} = 1 - JSD(Q_{purpose}||P_{purpose}) \quad (3)$$

3. Activity Interval Score: Measures the similarity between the distribution of time spent between trips (i.e., activity duration) in the synthetic diary and that of its cohort using JSD. To measure the distributions, durations are categorized into distinct bins: <15 min, 15-30 min, 30-60 min, 1-2 hrs, 2-4 hrs, and >4 hrs.

$$\text{Activity Interval Score} = 1 - JSD(Q_{interval}||P_{interval}) \quad (4)$$

4. Mode Distribution Score: Measures the similarity between the synthetic diary's mode distribution and the cohort's mode distribution using the JSD.

$$\text{Mode Distribution Score} = 1 - JSD(Q_{mode}||P_{mode}) \quad (5)$$

Finally, the Overall Realism Score is computed as ¼ (Trip Score + Purpose Score + Activity Interval Score + Mode Score) to provide a balanced diary quality assessment. Equal weighting of each component was chosen for simplicity and interpretability. Equal weighting was selected for simplicity and interpretability; Section 5.3 evaluates the sensitivity of the model comparison to four alternative component-weighting schemes.

*4.2. Aggregate-Level validation*

As a second validation, an aggregate-level analysis was conducted where the distributions of synthetic diaries are compared to the distribution of the entire HTS population. For the scores of Purpose Distribution, Activity Interval, and Mode distribution, the formulae remain the same as the one-to-cohort analysis. However, the sets $Q$ and $P$ instead comprise the global set of generated and HTS diaries, respectively, rather than a single entry and its cohort. Since trip count in the aggregate analysis is a distribution across all personas rather than a single true value, the Trip Count Score is also calculated using JSD to measure the similarity between the two sets of distributions:

$$\text{Trip Count Score}_{agg} = 1 - JSD(Q_{trips}||P_{trips}) \quad (6)$$

Including aggregate validation provides a macro-level analysis complementary to the micro-level one-to-cohort approach. The aggregate Overall Realism Score was calculated using the same equal four-component weighting as the one-to-cohort score: ¼ (Trip Score + Purpose Score + Activity Interval Score + Mode Score). Note that this is not a test of population synthesis.

We are not evaluating the distribution of the synthetic personas themselves but instead evaluating whether the travel pattern generated for this set of personas is, overall, representative of the entire HTS population's travel patterns.

### 4.3. The classical benchmark model

To establish a baseline, a benchmark using classical travel demand models was developed. The two-phase calibration–generation process first calibrates models on HTS data (weighted by population) with block-group land-use and demographic information, estimating coefficients to capture statistical relationships between person attributes (age, vehicle count, population and employment density, etc.) and observed travel behavior.

The benchmark models were selected as transparent, calibration-based references rather than as an exhaustive representation of available travel-behavior models. Daily trip frequency was predicted using negative binomial regression. This specification was selected because daily trip frequency is a nonnegative count variable and the HTS calibration data exhibit overdispersion: the observed mean number of daily trips is 3.72, whereas the variance is 7.33. Negative binomial regression relaxes the equidispersion assumption of the Poisson model and is therefore appropriate for these data (Jang, 2005). Multinomial logit models were used for travel mode and trip purpose because both are mutually exclusive categorical outcomes and these models provide established and interpretable probabilistic benchmarks for transportation applications (Anas, 1981; Penn et al., 2008). The negative binomial and mode-choice models were implemented using statsmodels, whereas the multinomial trip-purpose model was implemented using scikit-learn (Pedregosa et al., 2011; Seabold and Perktold, 2010).

Although the benchmark models require HTS calibration and therefore are not proposed as a solution for data-sparse applications, their application in Connecticut provides a familiar reference for comparing the zero-shot LLM with a conventional, survey-calibrated modeling workflow. More flexible alternatives include hurdle or zero-inflated models for count outcomes, mixed or nested logit models for categorical choices, and machine-learning methods capable of representing nonlinear relationships. Previous transportation research has found that machine-learning classifiers can outperform MNL in some mode-choice applications (Hagenauer and Helbich, 2017). These alternatives were not included because the objective of this study was to evaluate the viability of zero-shot LLM-based diary generation relative to an established and reproducible reference, rather than to identify the best predictive model for each travel component. Consequently, the findings are limited to the selected classical benchmarks and should not be interpreted as evidence that the LLM would outperform all advanced statistical or machine-learning models.

Next, these calibrated models were used to generate a new set of travel diaries. This generation process is run for the same set of synthetic personas that were provided to the LLM. This research does not focus on population synthesis; instead, the use of a shared set of personas isolates the evaluation to the task of activity generation alone. This ensures an "apples-to-apples" comparison, with LLM and classical benchmark starting from identical demographics and producing the same output format, enabling evaluation.

## 5. Results and discussion

This section presents the outcomes of the validation experiments, comparing the performance of the LLM-based diary generation framework against a benchmark of classical models through one-to-cohort and aggregate analyses. Additionally, it evaluates model robustness by examining component-weight sensitivity, multi-run stability, and concludes with illustrative examples of LLM-generated travel diaries.

### *5.1. One-to-Cohort validation performance*

Model performance is measured using one-to-cohort overall realism scores, with the results for the LLM and classical benchmark summarized in Table 2. The LLM-based approach achieved a mean score of 0.692, compared with 0.628 for the classical benchmark. Figure 6 shows an approximately unimodal distribution for the LLM scores, whereas the classical-benchmark distribution includes a secondary mode at a lower score. Because both models were evaluated for the same 2,143 synthetic personas, a paired t-test was used. The mean paired difference was 0.064 (95% confidence interval: 0.058–0.069), and the difference was statistically significant, t(2,142) = 24.02, ($P < 0.001$).

The difference in the means and score distributions indicates that the LLM-generated diaries achieved higher average similarity to their matched HTS cohorts under the proposed Overall Realism Score. Another notable result is that the standard deviation of the LLM scores (0.074) was lower than that of the classical benchmark (0.115). The distributions shown in Figure 6 further illustrate this difference. The LLM score distribution is narrower and more concentrated around its mean, indicating lower between-persona variability and fewer low-scoring outputs under the proposed metric. In contrast, the classical-benchmark distribution is wider and contains a secondary mode at a lower realism score of approximately 0.35. Overall, the LLM produced higher and less variable Overall Realism Scores across the 2,143 evaluated personas.

To further examine the strengths and limitations of each model, we analyzed their cohort-based performance across different demographic and behavioral subgroups. Figure 7 presents the mean realism scores for both models, segmented by employment status and age. A disaggregate analysis of model performance reveals that the LLM outperforms the classical benchmark across all demographic groups, including common traveler archetypes (middle-aged, employed). This advantage is also seen when modeling the less predictable segments, such as the unemployed and older adults. These groups are often difficult to capture in traditional models, as their travel behaviors are not anchored by the predictable, daily work trips characteristic of a typical full-time employee (Rasouli et al., 2015).

**Table 2**
Overall realism score summary for the LLM model versus the classical benchmark.

| Metric | Classical Benchmark | LLM Model |
|---|---|---|
| Mean Score | 0.628 | 0.692 |
| Median Score | 0.648 | 0.699 |
| Standard Deviation | 0.115 | 0.074 |
| Number of Personas | 2143 | 2143 |

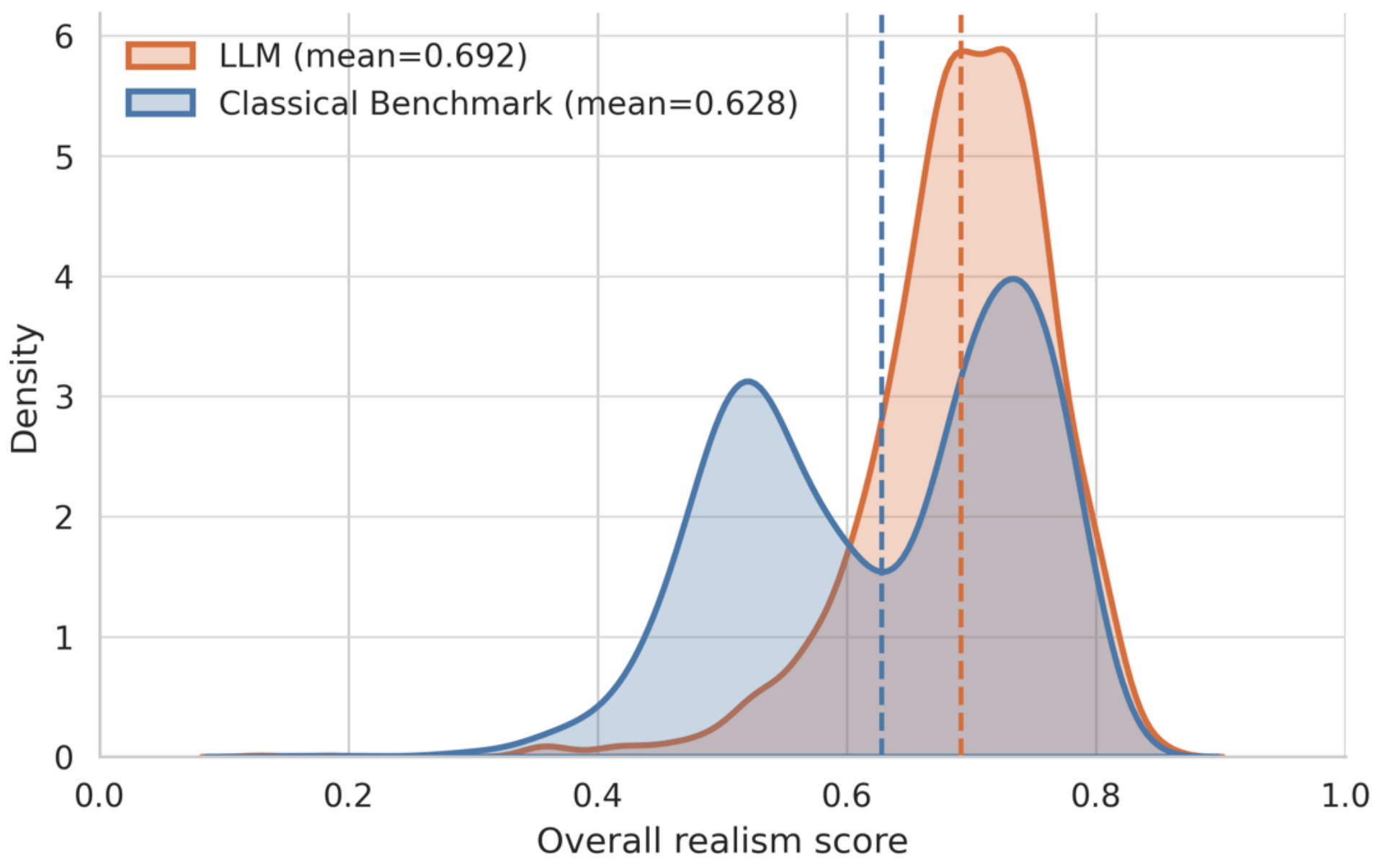

**Fig. 6.** The distribution of overall realism scores.

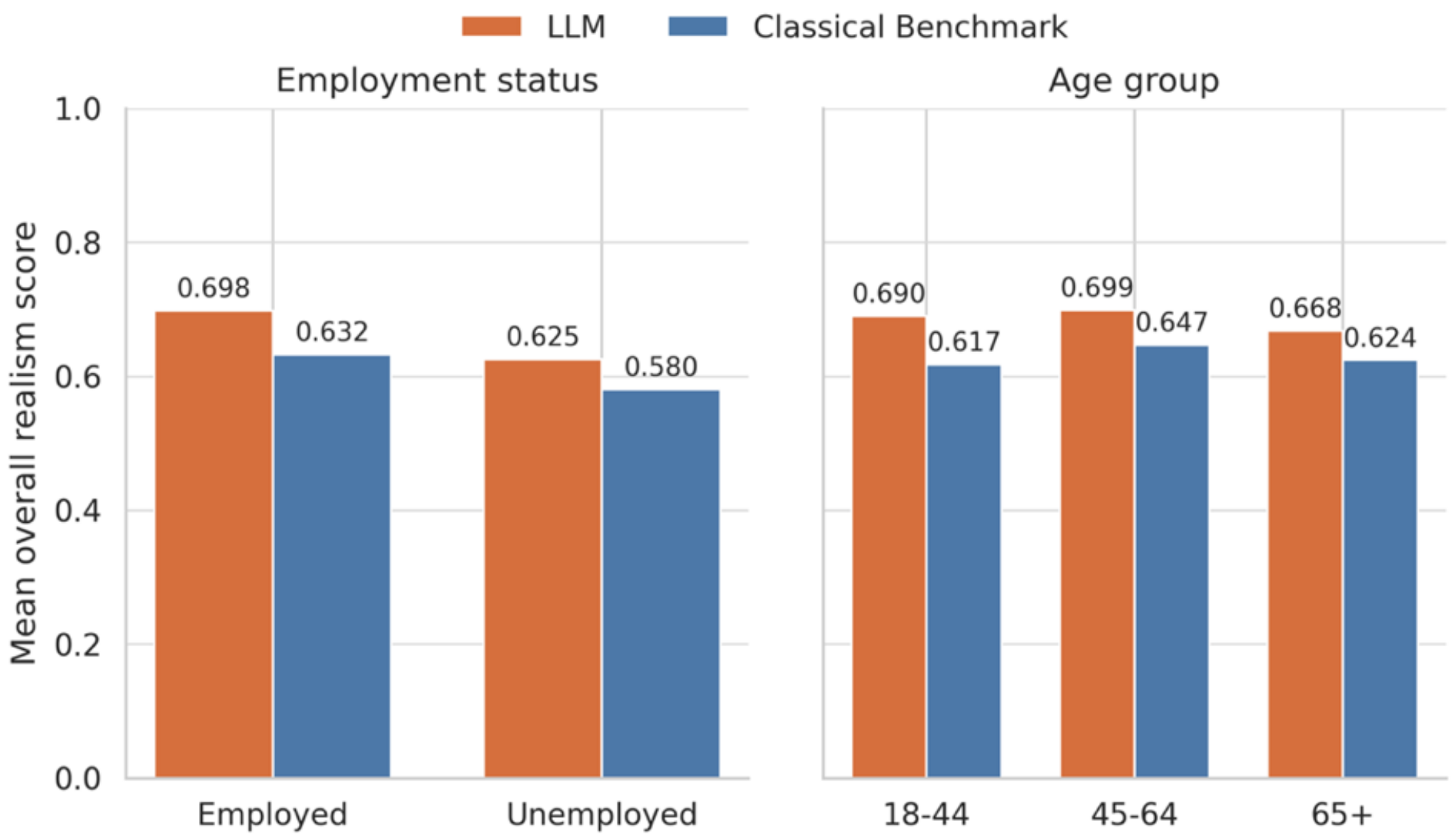

**Fig. 7.** Comparative cohort performance by demographic subgroups.

Figure 8 presents the distributions of one-to-cohort Overall Realism Scores by the final achieved cohort-matching level. Under Broad matching, the median score was 0.704 for the LLM and 0.496 for the classical benchmark, with interquartile ranges (IQRs) of 0.107 and 0.052, respectively. Under Strict matching, the corresponding medians were 0.685 and 0.577, with IQRs of 0.098 and 0.168. Under Ultra-Strict matching, the medians were 0.705 and 0.698, with IQRs of 0.082 and 0.186. Thus, the LLM had a higher median score at each achieved matching level. Its score distribution was narrower under Strict and Ultra-Strict matching but wider under Broad matching.

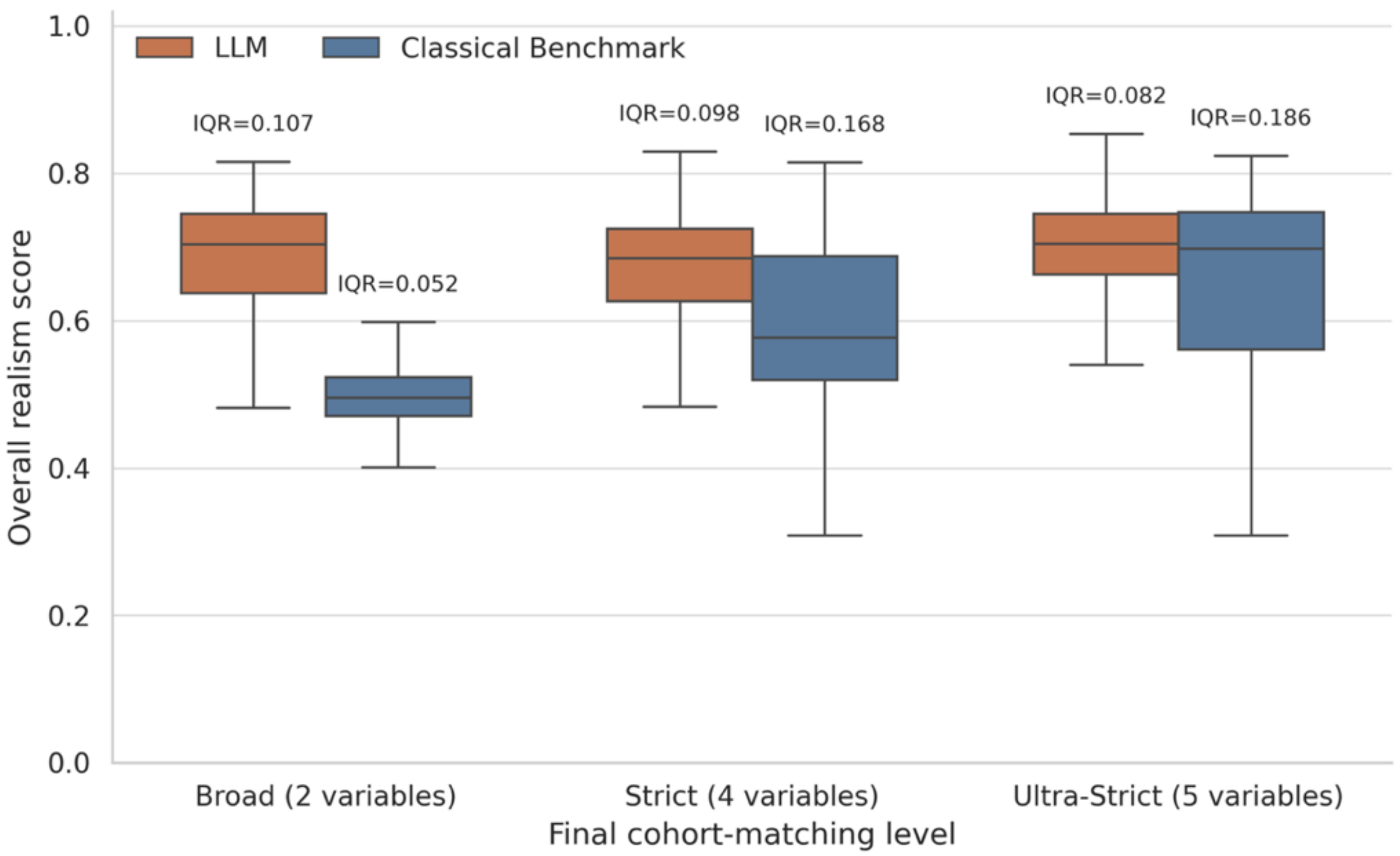

**Fig. 8.** Overall realism score distributions by matching level.

Lastly, Figure 9 provides a detailed, component-level breakdown of model performance. The analysis segments performance by the stringency of the cohort matching. The results reveal a clear trade-off: the classical benchmark excels in replicating numerical aspects of travel, outperforming trip count and interval scores. This strength is a direct result of using specialized statistical models, such as Negative Binomial, that were explicitly calibrated on the HTS dataset. This is not a shortcoming of the LLM, but rather an affirmation that traditional models are effective at fitting to known statistical distributions. The LLM, however, shows an advantage in the semantic task of assigning a trip purpose. This superiority is statistically significant ($p < 0.001$) across all cohort matching levels, highlighting its strength in capturing the "why" behind travel decisions. This trade-off is further proved by the mode choice results. The LLM perform more accurate when modeling complex or less common travel scenarios, maintaining a high degree of realism score.

The one-to-cohort validation highlights a trade-off in which classical models offer precise statistical fit for trip counts and durations, while LLMs better capture semantic aspects like trip purpose and travel mode without calibration.

### *5.2. Aggregate-Level validation performance*

In the aggregate-level validation, each model's full set of generated diaries was compared against the statistical distribution of the entire HTS population (Figure 10). The LLM achieved a substantially higher overall score than the classical benchmark (0.779 vs. 0.706), driven by its superior ability to replicate population-wide distributions of trip count, purpose, and mode.

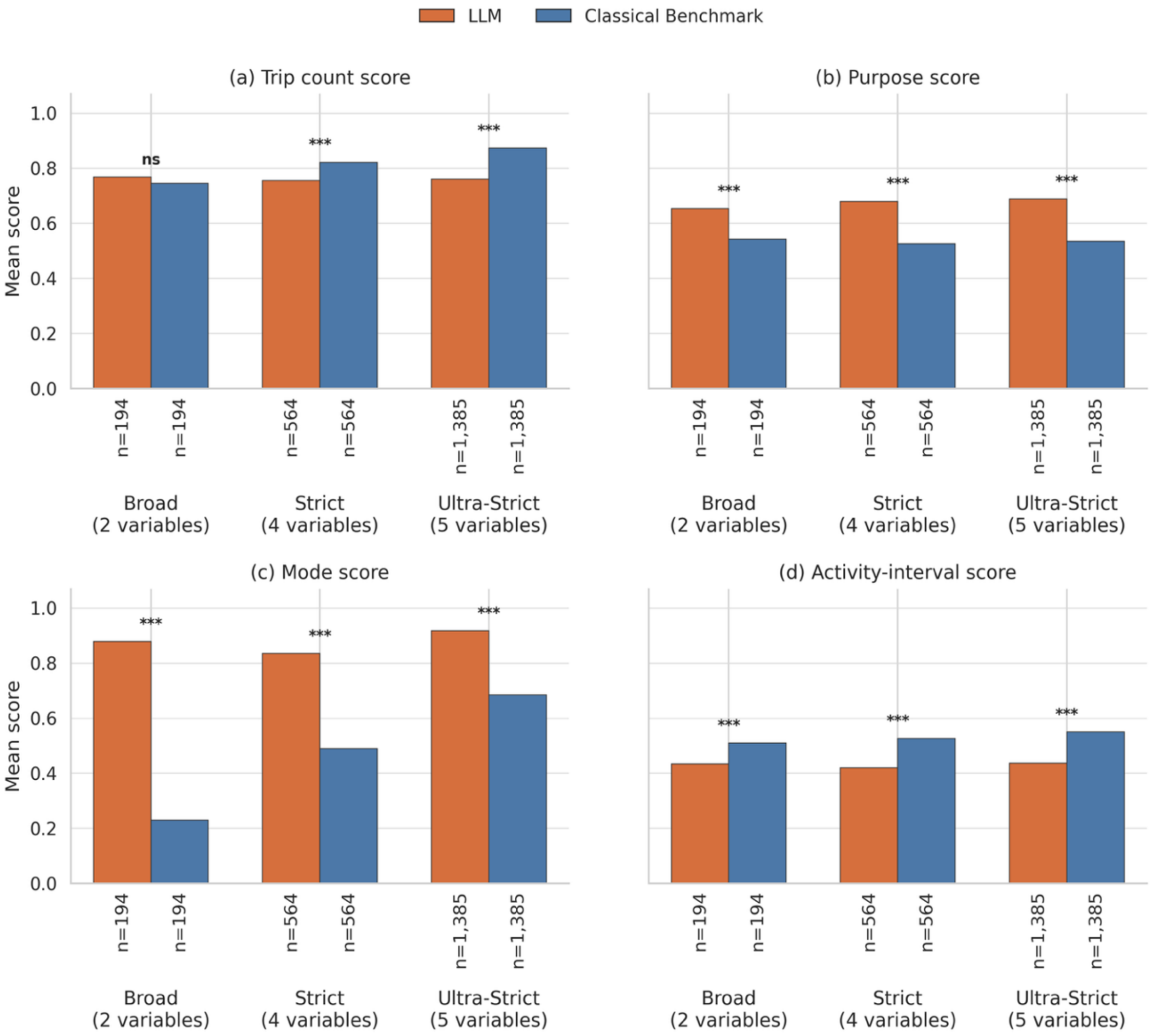

Paired t-test significance: ns, p≥0.05; * p<0.05; ** p<0.01; *** p<0.001. n = paired persona-days per model.

**Fig. 9.** Component score comparison by matching level. (a) trip count score, (b) purpose score, (c) mode score, (d), interval score. Significance values utilize paired t-test.

The classical benchmark's only advantage in the aggregate analysis was the activity-interval score, which may reflect its direct estimation from observed HTS travel data. The difference between the aggregate and one-to-cohort overall scores should, however, be interpreted in light of their distinct levels of comparison. In the one-to-cohort analysis, each generated diary is evaluated separately against a demographically matched HTS cohort. The trip-count component compares the generated diary's trip count with the matched cohort's mean, whereas the trip-purpose, travel-mode, and activity-interval components compare diary-level and cohort-level distributions using JSD. In the aggregate analysis, all generated diaries are pooled and compared with the entire HTS sample. Trip purpose, travel mode, and activity interval are evaluated using the same JSD formulation, but the comparison is between population-level distributions. Trip count is also represented as a population-level distribution and is therefore evaluated using JSD rather than as a deviation from a cohort mean. Thus, the aggregate overall score and the mean one-to-cohort score evaluate related but distinct aspects of realism and are not directly comparable numerical measures.

Pooling diaries reduces the sparsity of diary-level distributions and allows deviations across personas to offset one another in the resulting population-level distributions. Consequently, a model can reproduce broad population patterns even when some individual diaries differ from the behavior observed among their matched cohorts. The aggregate results indicate that the LLM reproduces the overall distributions of trip purpose and travel mode well, but they do not demonstrate that every generated diary is behaviorally consistent at the individual level. Although the LLM achieved a higher mean one-to-cohort overall realism score and a narrower score distribution than the classical benchmark, the component-level results identify remaining limitations in individual trip-frequency and activity-interval precision. Overall, these findings suggest that the current zero-shot framework is more effective at reproducing aggregate travel patterns than at consistently capturing all dimensions of travel behavior at the individual-diary level.

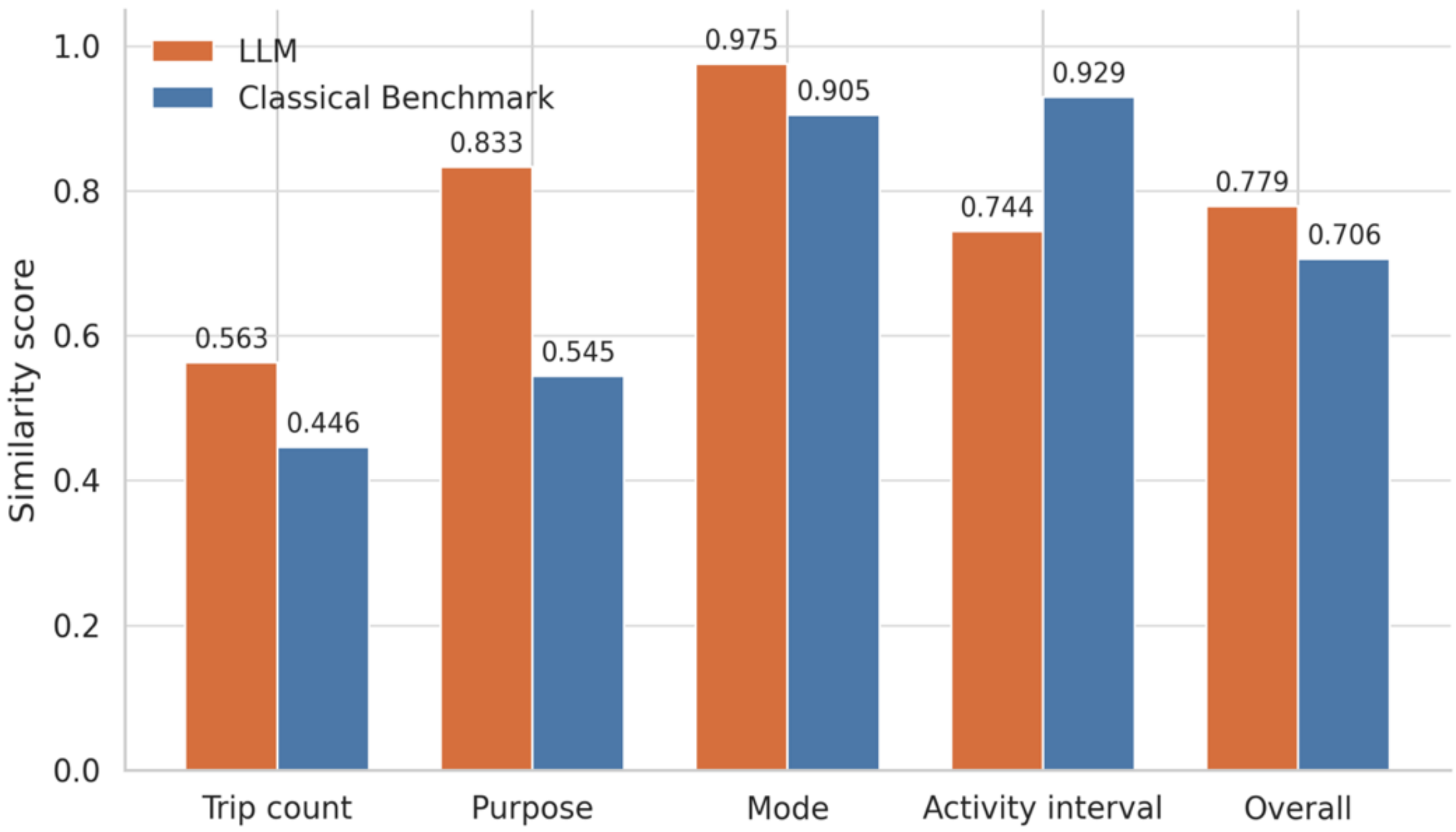

**Fig. 10.** Aggregate-level realism score components.

*5.3. Sensitivity to component weights*

The baseline overall realism score assigns equal weights (0.25) to trip count, trip purpose, travel mode, and activity interval. To examine whether the model comparison depends on this choice, four alternative weighting schemes were evaluated. In each alternative, one component received a higher weight of 0.40 and each remaining component received a lower weight of 0.20. The generated diaries and the four component scores were held unchanged. This analysis evaluates sensitivity to the construction of the composite score rather than the sensitivity of diary generation.

The LLM retained a higher mean overall realism score under all five weighting schemes at both validation levels as shown in Figure 11. In the one-to-cohort analysis, the difference between the LLM and classical benchmark ranged from 0.029 when activity interval received a greater weight to 0.111 when travel mode had a higher weight. At the aggregate level, the difference ranged from 0.021 under activity-interval emphasis to 0.116 under greater trip-purpose weighting. The smaller margin under greater activity-interval weighting is consistent with the classical benchmark's higher activity-interval score. Overall, the comparison between the LLM and classical benchmark remained consistent across the weighting schemes examined.

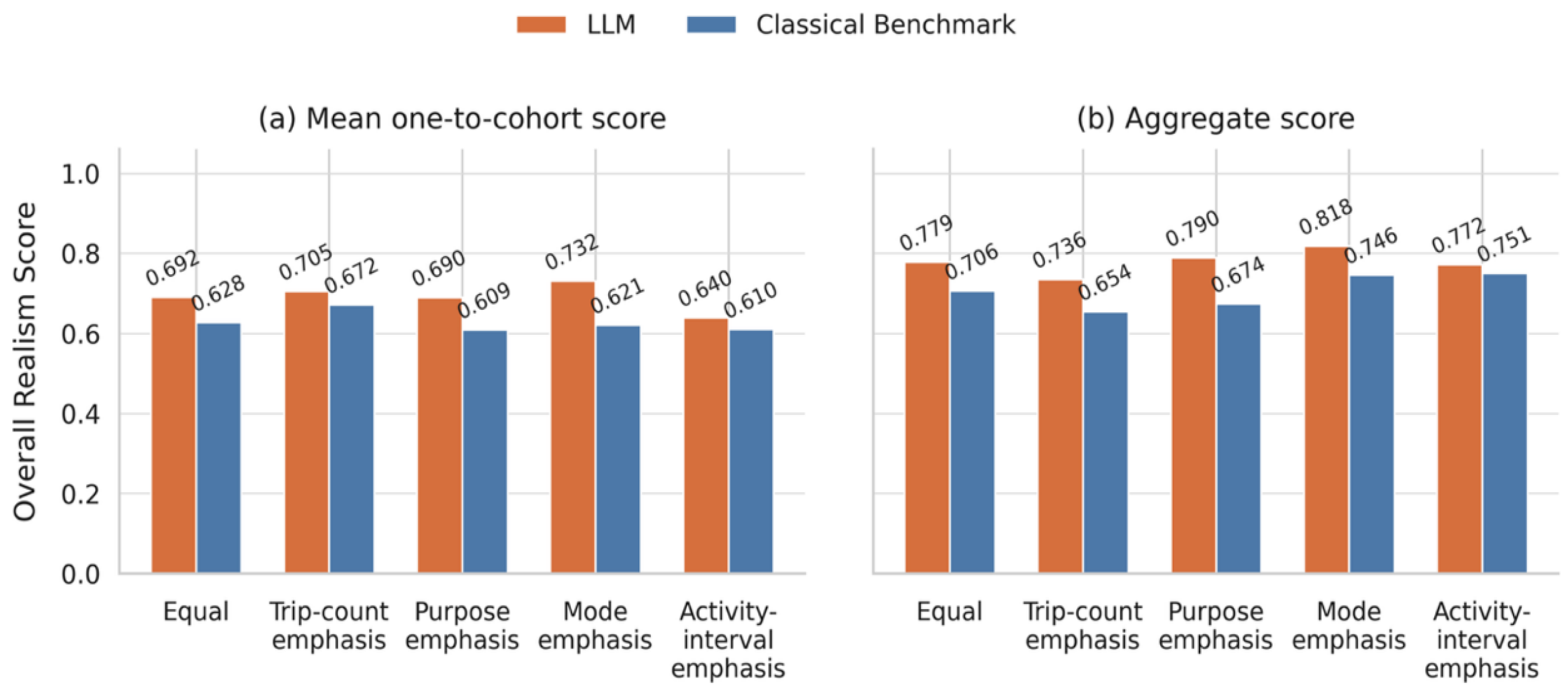

**Fig. 11.** Sensitivity of the one-to-cohort and aggregate Overall Realism Scores to alternative component weights.

*5.4. Run-to-run stability and illustrative outputs*

The complete LLM diary-generation framework was executed seven times to evaluate the stability of generated travel patterns across repeated runs. Each run processed the same 2,143 block groups and generated one persona-day diary per block group. Because the output dataset contains one record per generated trip rather than one row per diary, the number of trip records ranged from 6,813 to 6,931 across the seven runs. As shown in Table 3, mean trip frequency ranged from 3.179 to 3.234 trips per persona-day, with an across-run mean of 3.202 and a standard deviation of 0.018. Mean trip distance across all travel modes was also stable, with an across-run mean of 6.538 miles and a standard deviation of 0.045 miles.

Aggregate stability was evaluated by comparing every unique pair of the seven runs, resulting in 21 pairwise comparisons. For each pair, similarity was calculated as one minus the JSD for the trip-count, trip-purpose, travel-mode, and activity-interval distributions. Under this formulation, a value of 1 indicates identical aggregate distributions, while values closer to 1 indicate greater similarity.

As shown in Figure 12, the mean pairwise similarity was 0.9986 for trip count, 0.9924 for trip purpose, 0.9968 for travel mode, and 0.9994 for activity interval. The equal-weight overall similarity averaged 0.9968, with a standard deviation of 0.0004 and a range of 0.9960–0.9975. Thus, the seven executions did not reproduce identical individual travel diaries, but they generated highly consistent aggregate travel patterns. The individual diaries should therefore be interpreted as plausible stochastic realizations, whereas the aggregate distributions demonstrate strong run-to-run stability.

**Table 3**
Run-level stability of the LLM diary-generation framework across seven executions.

| Run | Persona-day diaries | Trip records | Mean trips per persona-day | Mean trip distance across all modes (miles) |
|---|---|---|---|---|
| Run 1 | 2,143 | 6,822 | 3.183 | 6.562 |
| Run 2 | 2,143 | 6,931 | 3.234 | 6.577 |
| Run 3 | 2,143 | 6,871 | 3.206 | 6.553 |
| Run 4 | 2,143 | 6,875 | 3.208 | 6.476 |
| Run 5 | 2,143 | 6,813 | 3.179 | 6.486 |
| Run 6 | 2,143 | 6,859 | 3.201 | 6.523 |
| Run 7 | 2,143 | 6,863 | 3.203 | 6.591 |
| Across-run mean | 2,143 | 6,862 | 3.202 | 6.538 |
| Across-run SD | 0.0 | 38.8 | 0.018 | 0.045 |

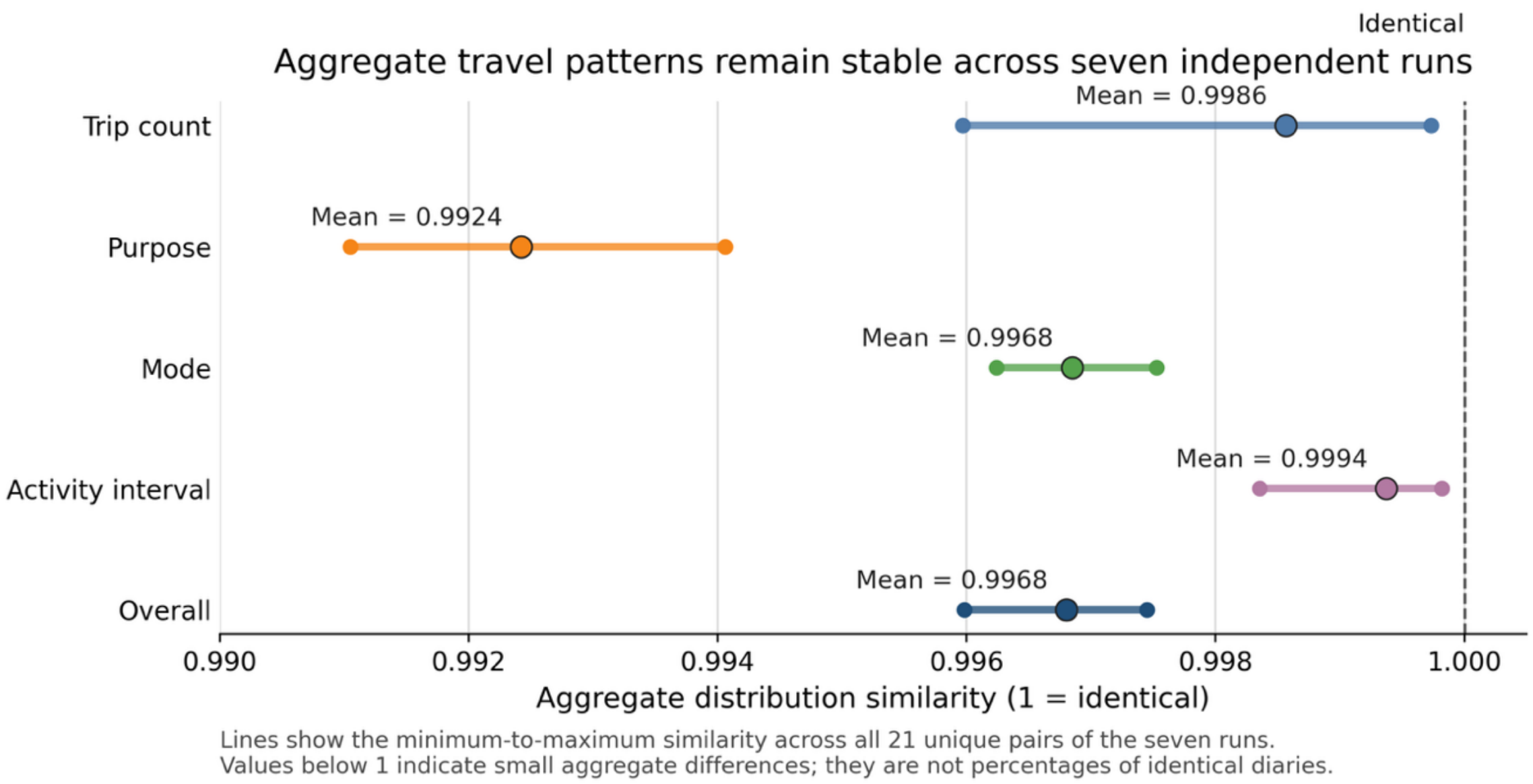

**Fig. 12.** Aggregate distributional stability across seven independent executions of the LLM diary-generation framework.

To complement the aggregate stability assessment, Table 4 presents three sample diaries from the primary LLM run. The examples represent low-, medium-, and high-walkability contexts. Start and end times indicate the travel interval for each trip, while the time between the end of one trip and the beginning of the next represents the implied duration of the destination activity. The low-walkability example contains a chronological workplace–errand–home sequence and implies 7 hours 45 minutes at the primary workplace. The medium-walkability example assigns only nonwork activities to an unemployed persona, including an errand, grocery shopping, and a return-home trip. The high-walkability example contains a workplace–errand–home sequence and implies 6 hours 40 minutes at the primary workplace. In the three displayed examples, the trips are chronologically ordered and nonoverlapping, end with a return-home trip, and are consistent with the personas' employment status and household vehicle availability. These examples provide a qualitative illustration of the generated diary structure; the quantitative findings are based on the complete generated dataset and the full validation analysis.

**Table 4**
Illustrative LLM-Generated Travel Diaries.

| Person | Trip | Travel interval | Trip purpose | Mode and distance |
|---|---|---|---|---|
| A | 1 | 7:30–8:15 AM | Go to primary workplace | Household vehicle; 10.1 mi |
| | 2 | 4:00–4:45 PM | Run errands | Household vehicle; 4.8 mi |
| | 3 | 6:15–7:00 PM | Go home | Household vehicle; 5.2 mi |
| B | 1 | 1:30–2:20 PM | Run errands | Household vehicle; 3.1 mi |
| | 2 | 3:10–4:05 PM | Grocery shopping | Household vehicle; 4.8 mi |
| | 3 | 5:00–5:45 PM | Go home | Household vehicle; 5.2 mi |
| C | 1 | 6:30–7:20 AM | Go to primary workplace | Household vehicle; 12.1 mi |
| | 2 | 2:00–2:50 PM | Run errands | Household vehicle; 4.8 mi |
| | 3 | 5:15–6:05 PM | Go home | Household vehicle; 12.1 mi |

*- (A) Low-walkability; Persona 1710 - GEOID 090091574002; NWI = 4.83; employed, age 45–54; upper-middle income; household size = 3; 3+ vehicles*
*- (B) Medium-walkability; Persona 1612 - GEOID 090093433001; NWI = 13.00; unemployed, age 35–44; upper-middle income; household size = 2; 1 vehicle*
*- (C) High-walkability; Persona 1565 - GEOID 090091702003; NWI = 14.83; employed, age 35–44; low income; household size = 2; 1 vehicle*
*- These examples were selected to illustrate internally coherent outputs rather than to represent all generated diaries. (NWI = National Walkability Index; HH = household).*

## 6. Conclusions

This study introduced an activity generation model that utilizes prompt-based LLM techniques to create individual-level travel diaries. By grounding persona and prompt generation in open-source census and land-use data, our approach avoids the traditional reliance on proprietary household travel surveys for model calibration. A key aspect of this work is its validation strategy, which moves beyond simple aggregate comparisons by evaluating the "realism" of generated diaries through a detailed one-to-cohort analysis against real household travel survey diaries. Results demonstrate that LLMs provide a powerful foundation for agent-based demand modeling.

The results reveal mixed performance of the LLM compared to a classical benchmark consisting of traditional models (e.g., Negative Binomial, MNL) calibrated on the HTS dataset. In the one-to-cohort analysis, the classical benchmark outperformed the LLM on trip count and trip interval, reflecting the strength of such methods in fitting to a distribution of known datasets. In contrast, the LLM demonstrated comparable mode choice scores and substantially outperformed the MNL method in assigning trip purpose. This highlights a key advantage of LLM's language-based reasoning in capturing the "why" behind travel behavior.

While the scores of individual components highlight trade-offs between numerical precision and the LLM's semantic reasoning, the overall realism scores from one-to-cohort analyses show that the LLM generated diaries are, on average, holistically more realistic than the classical models (scores of 0.692 and 0.628, respectively). Further, the LLM scores of overall realism were distributed narrowly, whereas the classical models indicated a bimodal distribution. Therefore, while classical methods can achieve high precision, they are at risk of producing low-scoring diaries, whereas the LLM maintains a more consistent overall realism.

Beyond the one-to-cohort analysis, an aggregate analysis was conducted to assess population-level travel patterns. The LLM achieved a higher aggregate overall realism score than the classical benchmark (0.779 versus 0.706). Because aggregate validation pools all generated diaries and evaluates population-level distributions, this result should not be interpreted as evidence that every LLM-generated diary is individually consistent. Rather, the combined results indicate that the LLM captures broad population-level trip-purpose and travel-mode distributions effectively, while the one-to-cohort component scores show limitations in individual trip-frequency and activity timing.

These results should be interpreted in relation to how the two approaches were developed. The classical benchmark models provide calibration-based references because their parameters were estimated directly from the HTS data, whereas the LLM was applied without study-specific training or calibration using those data. Accordingly, the findings show that the zero-shot LLM achieved comparable or higher realism scores relative to the selected classical benchmarks; they do not establish that the LLM would outperform all advanced travel-behavior models. This distinction is particularly relevant to the intended application of the framework in settings where sufficiently detailed local household travel-survey data are limited or unavailable.

Another consideration is whether the accuracy of the generated diaries varies across demographic and socioeconomic groups or geographic settings. Because Llama 3 was pretrained on large-scale publicly available online data, the model may represent some population groups and geographic settings better than others (Gallegos et al., 2024; Llama 3 model card - Github, 2024). This could result in systematic differences in generated trip rates, mode shares, trip purposes, or travel times across groups or areas. If carried into a transportation analysis, these differences could lead to over- or underestimation of travel demand and transit needs and could affect accessibility assessments and the estimated benefits of transportation investments. The generated diaries in this study were used to evaluate the proposed diary-generation approach, not to support actual planning or policy decisions. Before operational use, the framework should be evaluated against observed local travel data across relevant population groups and geographic areas (NIST, 2024). Future research should also examine a hybrid approach in which locally estimated travel relationships are used to calibrate or correct the LLM-generated outputs.

The present empirical evaluation is limited to Connecticut, and the findings should not be assumed to apply directly to other regions. Although the general framework can be implemented elsewhere, the ACS and SLD inputs must be replaced with data for the target area, and region-specific classifications, persona-synthesis parameters, prompt descriptions, and travel-mode and trip-purpose categories may need to be adjusted to reflect local demographic, land-use, and transportation condition. The generated travel diaries should also be compared with travel survey data from the study area, when available, to evaluate how well the framework represents local travel patterns.

The LLM prompt methodology could be enhanced by implementing a two-stage generation process where activity generation is separate from scheduling, forcing a set of activities to be allocated (or discarded) to fit within a 24-hour day. This would address the finding that the LLM was outperformed in interval score by providing an aspect of temporal realism. In addition, while this method was developed for zero-shot application, the realism score could be used as a feedback mechanism to train or iteratively guide the LLM in contexts where a validation set is available. Lastly, the framework generates individual travel diaries informed by household characteristics (e.g., size, vehicle availability) but does not model complex intra-household dependencies such as joint trips, which could be addressed in future multi-agent extensions.

Ultimately, this work demonstrates that LLMs are a promising complementary tool rather than an immediate replacement for established travel-demand modeling methods. Future research should compare the framework with more flexible statistical and machine-learning benchmarks (e.g., hurdle or zero-inflated count models, mixed or nested logit models, random forests, gradient-boosted trees, and sequence-based models) using held-out HTS observations. Future work should also investigate a hybrid framework in which statistical or machine-learning components calibrated using observed local travel data provide constraints on trip frequency, timing, and mode feasibility, while the LLM generates contextually coherent activity and trip-purpose sequences.

## Funding

This research did not receive any specific grant from funding agencies in the public, commercial, or not-for-profit sectors.

## Declaration of Competing Interest

The authors declare that they have no known competing financial interests or personal relationships that could have appeared to influence the work reported in this paper.

## Acknowledgments

The authors declare that the contents of this article have not been published previously. The authors confirm contribution to the paper as follows: study conception and design: S. Golrokh Amin, D. Rhoads, F. Fakhrmoosavi; data processing: S. Golrokh Amin, D. Rhoads, N. Lownes; computational modeling: S. Golrokh Amin, D. Rhoads; analysis and interpretation of results: S. Golrokh Amin, D. Rhoads, F. Fakhrmoosavi, N. Lownes, J. Ivan; draft manuscript preparation: S. Golrokh Amin, D. Rhoads, F. Fakhrmoosavi, N. Lownes, J. Ivan. All authors reviewed the results and approved the final version of the manuscript.

## References

Ackley, D.H., Hinton, G.E., Sejnowski, T.J., 1985. A Learning Algorithm for Boltzmann Machines. Cognitive Science 9, 147–169. https://doi.org/10.1207/s15516709cog0901_7

Anas, A., 1981. The Estimation of Multinomial Logit Models of Joint Location and Travel Mode Choice from Aggregated Data. Journal of Regional Science 21, 223–242. https://doi.org/10.1111/j.1467-9787.1981.tb00696.x

Beneduce, C., Lepri, B., Luca, M., 2025. Large Language Models are Zero-Shot Next Location Predictors. IEEE Access 13, 77456–77467. https://doi.org/10.1109/ACCESS.2025.3565297

Bernardin, V.L., Trevino, S., Slater, G., Gliebe, J., 2015. Simultaneous Travel Model Estimation from Survey Data and Traffic Counts. Transportation Research Record: Journal of the Transportation Research Board 2494, 69–76. https://doi.org/10.3141/2494-08

Dargay, J., Gately, D., 1999. Income's effect on car and vehicle ownership, worldwide: 1960–2015. Transportation Research Part A: Policy and Practice 33, 101–138. https://doi.org/10.1016/S0965-8564(98)00026-3

Devroye, L., 2006. Nonuniform random variate generation, in: Handbooks in Operations Research and Management Science. Elsevier, pp. 83–121. https://doi.org/10.1016/S0927-0507(06)13004-2

Efron, B., Narasimhan, B., 2020. The Automatic Construction of Bootstrap Confidence Intervals. Journal of Computational and Graphical Statistics 29, 608–619. https://doi.org/10.1080/10618600.2020.1714633

Fakhrmoosavi, F., Amin, S.G., He, S., 2025. Impacts of Shared Autonomous Vehicles on Traffic Operations and Infrastructure Durability in Connecticut. Transportation Infrastructure Durability Center (TIDC).

Fakhrmoosavi, F., Paithankar, P., Kockelman, K.M., Huang, Y., Hawkins, J., 2024. Self-driving vehicles' impacts on Americans' long-distance domestic travel choices. Transportation Planning and Technology 47, 1262–1276. https://doi.org/10.1080/03081060.2023.2288629

Gallegos, I.O., Rossi, R.A., Barrow, J., Tanjim, M.M., Kim, S., Dernoncourt, F., Yu, T., Zhang, R., Ahmed, N.K., 2024. Bias and fairness in large language models: A survey. Computational linguistics 50, 1097–1179. https://doi.org/10.1162/coli_a_00524

Giuliano, G., Dargay, J., 2006. Car ownership, travel and land use: a comparison of the US and Great Britain. Transportation Research Part A: policy and practice 40, 106–124. https://doi.org/10.1016/j.tra.2005.03.002

Guan, X., Huang, S., Chen, C., 2025. Using Multiple Biased Data Sets to Recover Missing Trips with a Behaviorally Informed Model. Transportation Science trsc.2024.0550. https://doi.org/10.1287/trsc.2024.0550

Hagenauer, J., Helbich, M., 2017. A comparative study of machine learning classifiers for modeling travel mode choice. Expert Systems with Applications 78, 273–282. https://doi.org/10.1016/j.eswa.2017.01.057

Holian, M.J., 2020. The impact of urban form on vehicle ownership. Economics letters 186, 108763. https://doi.org/10.1016/j.econlet.2019.108763

Holtzman, A., Buys, J., Du, L., Forbes, M., Choi, Y., 2020. The Curious Case of Neural Text Degeneration. https://doi.org/10.48550/arXiv.1904.09751

Hörl, S., Balać, M., 2020. Open data travel demand synthesis for agent-based transport simulation: A case study of Paris and Île-de-France. Reproducible scenarios for agent-based transport simulation: A case study for Paris and Île-de-France 1499. https://doi.org/10.3929/ethz-b-000412979

Jang, T.Y., 2005. Count Data Models for Trip Generation. J. Transp. Eng. 131, 444–450. https://doi.org/10.1061/(ASCE)0733-947X(2005)131:6(444)

Kim, T., Sharda, S., Zhou, X., Pendyala, R.M., 2020. A stepwise interpretable machine learning framework using linear regression (LR) and long short-term memory (LSTM): City-wide demand-side prediction of yellow taxi and for-hire vehicle (FHV) service. Transportation Research Part C: Emerging Technologies 120, 102786. https://doi.org/10.1016/j.trc.2020.102786

Liang, Y., Ding, F., Huang, G., Zhao, Z., 2023. Deep trip generation with graph neural networks for bike sharing system expansion. Transportation Research Part C: Emerging Technologies 154, 104241. https://doi.org/10.1016/j.trc.2023.104241

Lim, H., Uddin, M., Liu, Y., Chin, S.-M., Hwang, H.-L., 2022. A Comparative Study of Machine Learning Algorithms for Industry-Specific Freight Generation Model. Sustainability 14, 15367. https://doi.org/10.3390/su142215367

Liu, T., Yang, J., Yin, Y., 2025. Toward LLM-Agent-Based Modeling of Transportation Systems: A Conceptual Framework. https://doi.org/10.48550/arXiv.2412.06681

Llama 3 model card - Github, 2024. Llama 3 model card [WWW Document]. Llama 3 model card. URL https://github.com/meta-llama/llama3/blob/main/MODEL_CARD.md

Meta, 2025. Ollama [WWW Document]. URL https://ollama.com (accessed 10.29.25).

Mladenovic, M., Trifunovic, A., 2014. The Shortcomings of the Conventional Four Step Travel Demand Forecasting Process. Journal of Road and Traffic Engineering.

Mori, K., Fakhrmoosavi, F., Gurumurthy, K.M., Camargo, P., Kockelman, K.M., 2026. Impacts of spatial resolution on agent-based transportation simulations with shared autonomous vehicles. Computers,

Environment and Urban Systems 124, 102371. https://doi.org/10.1016/j.compenvurbsys.2025.102371

NIST, 2024. Artificial intelligence risk management framework: Generative artificial intelligence profile. NIST Trustworthy and Responsible AI Gaithersburg. https://doi.org/10.6028/NIST.AI.600-1

Paithankar, P., Fakhrmoosavi, F., Kockelman, K.M., Perrine, K.A., 2024. International travel patterns: exploring destination preferences and airfare trends to and from the USA. Transportation Planning and Technology 47, 1243–1261. https://doi.org/10.1080/03081060.2024.2366241

Pedregosa, F., Varoquaux, G., Gramfort, A., Michel, V., Thirion, B., Grisel, O., Blondel, M., Prettenhofer, P., Weiss, R., Dubourg, V., others, 2011. Scikit-learn: Machine learning in Python. the Journal of machine Learning research 12, 2825–2830. https://doi.org/10.5555/1953048.2078195

Penn, M., Vargas, F., Chimba, D., 2008. Multinomial Modeling of Purpose Driven Trip, in: Transportation Land Use, Planning, and Air Quality. Presented at the Transportation Land Use, Planning, and Air Quality Congress 2007, American Society of Civil Engineers, Orlando, Florida, United States, pp. 64–78. https://doi.org/10.1061/40960(320)8

Pourebrahim, N., Sultana, S., Niakanlahiji, A., Thill, J.-C., 2019. Trip distribution modeling with Twitter data. Computers, Environment and Urban Systems 77, 101354. https://doi.org/10.1016/j.compenvurbsys.2019.101354

Püschel, J., Barthelmes, L., Kagerbauer, M., Vortisch, P., 2024. Comparison of discrete choice and machine learning models for simultaneous modeling of mobility tool ownership in agent-based travel demand models. Transportation Research Record 2678, 376–390. https://doi.org/10.1177/03611981231206175

Rasouli, T., Timmermans, H., Peter, van der W., 2015. Employment status transitions and shifts in daily activity-travel behavior with special focus on shopping duration. Transportation 42, 919–931. https://doi.org/10.1007/s11116-015-9655-5

Rezvany, N., Kukic, M., Bierlaire, M., 2024. A Review of Activity-based Disaggregate Travel Demand Models. Findings. https://doi.org/10.32866/001c.125431

Rostami, A., Kamjoo, E., Fakhrmoosavi, F., Zockaie, A., 2024. Estimating Path Travel Costs in Large-Scale Networks Using Machine-Learning Techniques. Transportation Research Record: Journal of the Transportation Research Board 2678, 57–75. https://doi.org/10.1177/03611981231172961

Sason, I., 2022. Divergence Measures: Mathematical Foundations and Applications in Information-Theoretic and Statistical Problems. Entropy 24, 712. https://doi.org/10.3390/e24050712

Seabold, S., Perktold, J., 2010. Statsmodels: econometric and statistical modeling with python., in: SciPy. Presented at the Proceedings of the 9th Python in Science Conference, pp. 92--96. https://doi.org/10.25080/Majora-92bf1922-011

Taş, M.B.H., Özkan, K., Sarıçiçek, İ., Yazici, A., 2025. Transportation Mode Selection Using Reinforcement Learning in Simulation of Urban Mobility. Applied Sciences 15, 806. https://doi.org/10.3390/app15020806

U.S. Bureau of Labor Statistics, 2015. Employment status of the civilian noninstitutional population by age, sex, and race [WWW Document]. URL https://www.bls.gov/cps/aa2015/cpsaat03.htm

Virtanen et al., 2020. SciPy 1.0: fundamental algorithms for scientific computing in Python. Nature methods 17, 261--272.

Wang, J., Jiang, R., Yang, C., Wu, Z., Onizuka, M., Shibasaki, R., Koshizuka, N., Xiao, C., 2024. Large Language Models as Urban Residents: An LLM Agent Framework for Personal Mobility Generation. https://doi.org/10.48550/arXiv.2402.14744

Zhang, F., Wu, T.-Y., Wang, Y., Xiong, R., Ding, G., Mei, P., Liu, L., 2020. Application of Quantum Genetic Optimization of LVQ Neural Network in Smart City Traffic Network Prediction. IEEE Access 8, 104555–104564. https://doi.org/10.1109/ACCESS.2020.2999608

Zhuang, Y., Azfar, T., Wang, Y., Sun, W., Wang, X., Guo, Q., Ke, R., 2024. Quantum Computing in Intelligent Transportation Systems: A Survey. CHAIN 1, 138–149. https://doi.org/10.23919/CHAIN.2024.000007